\documentclass[letterpaper, 10 pt, conference]{ieeeconf}  
\pdfoutput=1

\IEEEoverridecommandlockouts       

\usepackage[backend=biber,
            url=false,
            isbn=false,
            doi=false,
            backref=false,
            style=ieee,
            natbib=true,
            mincitenames=1,
            maxcitenames=1,
            citestyle=numeric-comp,
            sorting=nyt,
            block=none]{biblatex}

\usepackage{algorithm}
\usepackage[noend]{algpseudocode}

\usepackage{graphics}
\usepackage{bbm}
\usepackage[pdftex]{graphicx}
\usepackage{wrapfig}
\DeclareGraphicsExtensions{.pdf,.png,.jpg}
\usepackage{epsfig}
\usepackage[font={small}]{caption}
\usepackage{subcaption}
\usepackage[rightcaption]{sidecap}
\usepackage{pbox}

\usepackage{bigstrut}
\usepackage{mathtools}
\usepackage{amsmath, amssymb, amscd}
\usepackage{ wasysym } 
\usepackage{amsfonts}
\usepackage{mathptmx} 
\usepackage{gensymb} 
\usepackage{nicefrac}       
\numberwithin{equation}{section} 

\DeclareMathAlphabet{\mathcal}{OMS}{lmsy}{m}{n}
\DeclareSymbolFont{largesymbols}{OMX}{cmex}{m}{n}
\usepackage{textcomp} 

\usepackage{algorithm} 
\usepackage{algorithmicx, algpseudocode} 

\usepackage{array} 
\usepackage{tabularx}
\usepackage{multirow}
\usepackage{booktabs}
\usepackage{tabulary}

\usepackage[T1]{fontenc} 
\usepackage[utf8]{inputenc}
\usepackage[english]{babel} 
\usepackage{units}
\usepackage{bm}
\usepackage{times} 
\usepackage{xspace}
\usepackage{flushend}
\usepackage{balance} 
\usepackage{csquotes}
\usepackage{makeidx}
\usepackage{blindtext}

\let\labelindent\relax
\usepackage{enumitem}

\usepackage{ragged2e}
\usepackage{soul} 
\usepackage{subfiles} 

\usepackage[protrusion=true,expansion=true]{microtype}
\usepackage[yyyymmdd]{datetime}

\date{\protect\formatdate{1}{1}{2001}}

\usepackage{url}
\makeatletter
\g@addto@macro{\UrlBreaks}{\UrlOrds}
\makeatother
\usepackage{color}
\usepackage[usenames,dvipsnames,table,xcdraw]{xcolor}
\usepackage{pgfplots} 
\pgfplotsset{compat=newest}

\usepackage{marginnote}
\usepackage{soul} 

\usepackage[colorinlistoftodos]{todonotes}
\usepackage{multicol}
\newcommand{\tocite}[1]{%
\textcolor{red}{[cite:\ifthenelse{\equal{#1}{}}{}{#1}?]}
}

\newcommand{\ignore}[1]{}

\renewcommand{\baselinestretch}{1.0}



\title{\LARGE \bf
MMGSD: Multi-Modal Gaussian Shape Descriptors for Correspondence Matching in 1D and 2D Deformable Objects
}

\author{Aditya Ganapathi$^{*1}$, Priya Sundaresan$^{*1}$, Brijen Thananjeyan$^{1}$, Ashwin Balakrishna$^{1}$, \\ Daniel Seita$^{1}$, Ryan Hoque$^{1}$, Joseph E. Gonzalez$^{1}$, Ken Goldberg$^{1}$
\thanks{$^{*}$ Authors have contributed equally and names are in alphabetical order.}%
\thanks{$^{1}$University of California, Berkeley, USA}
\thanks{Correspondence to Aditya Ganapathi: \texttt{avganapathi@berkeley.edu}}
}

\begin{document}

\maketitle

\begin{abstract}
We explore learning pixelwise correspondences between images of deformable objects in different configurations. Traditional correspondence matching approaches such as SIFT, SURF, and ORB can fail to provide sufficient contextual information for fine-grained manipulation. We propose Multi-Modal Gaussian Shape Descriptor (MMGSD), a new visual representation of deformable objects which extends ideas from dense object descriptors to predict all symmetric correspondences between different object configurations. MMGSD is learned in a self-supervised manner from synthetic data and produces correspondence heatmaps with measurable uncertainty. In simulation, experiments suggest that MMGSD can achieve an RMSE of 32.4 and 31.3 for square cloth and braided synthetic nylon rope respectively. The results demonstrate an average of 47.7\% improvement over a provided baseline based on contrastive learning, symmetric pixel-wise contrastive loss (SPCL), as opposed to MMGSD which enforces distributional continuity. 
\end{abstract}


\section{Introduction}
\label{sec:introduction}

Robotic manipulation of deformable objects is a growing area of research that has exposed many limitations in perception systems~\cite{priya-rope,vision-based-rope-manip,seita_ryan,fabric_vsf_2020,thananjeyan2017multilateral,ganapathi2020learning, qian2020cloth, lerrel, chi2019occlusion, mcconachie2018estimating}. Acquiring useful visual state representations of deformable objects for manipulation is a central focus in prior work~\cite{sim2real_deform_2018, yan2020self, wang2019learning, priya-rope, schulman2016learning, tang2017state}. Inferring such representations is challenging due to the infinite dimensional state space, tendency to self-occlude, and often textureless and symmetric nature of deformable objects. One successful prior approach is learning pixelwise correspondences between images of deformable objects in different configurations as in~\citet{florence_2020, priya-rope,dense-obj-nets,ganapathi2020learning, schulman2013tracking, tang2017state, javdani2011modeling}. However, these methods
can fail to address uncertainty and symmetries, which can cause issues for downstream planning and control. Consider a robotic agent attempting to identify a corner of towel to fold it according to a video demonstration. The agent could leverage the fabric's inherent symmetry to manipulate the corner of the towel for which its uncertainty in its location is the lowest, since all four corners are viable options.
To enable such behaviors, we extend the correspondence learning algorithms from~\cite{priya-rope,dense-obj-nets,ganapathi2020learning} to (1) provide measures of uncertainty in predicted correspondences by formulating a distribution matching objective for correspondence learning inspired by~\cite{florence_2020} and (2) explicitly predicting symmetric correspondences. Experiments suggest that the learned correspondences for both 1D and 2D deformable objects are more stable and continuous than those used in prior work and are less prone to symmetrical ambiguities and provide uncertainty estimates. 




\begin{figure}[t]
\center
\includegraphics[width=1\columnwidth]{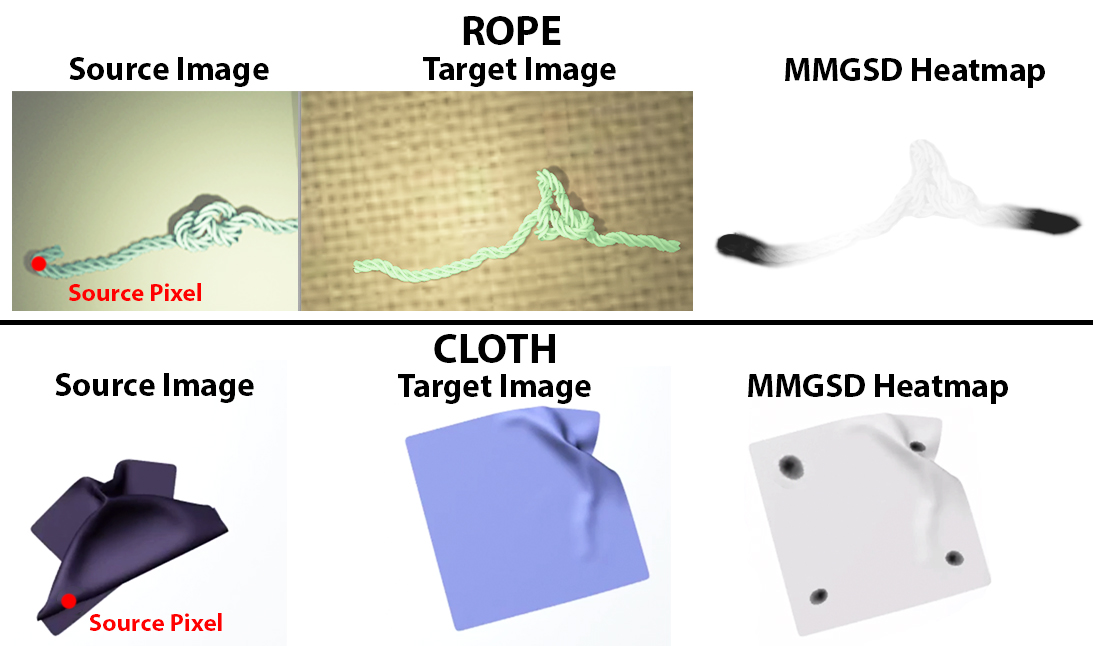}
\caption{Multi-Modal Gaussian Shape Descriptors (MMGSD) learns symmetry-aware pixelwise correspondences for multi-modal semantic keypoints between source and target images of deformable objects. We visualize the results of 2-modal and 4-modal MMGSD correspondence heatmaps for rope and cloth, respectively, relative to the source pixels from column 1.}
\label{fig:teaser}
\end{figure}

\section{Problem Statement}
\label{sec:problem_definition}

Given two images, $I_a$ and $I_b$, of a deformable object in two different configurations respectively, and a source pixel location $(u_a, v_a)$ (such that the pixel is $I_a[u_a,v_a]$), find its $n(u_a,v_a)$ pixel correspondences $\left((u_{b_i}, v_{b_i})\right)_{i=1}^{n(u_a,v_a)}$ in $I_b$. There may be multiple possible matches due to symmetry, such as when matching a corner of a square cloth in $I_a$ to all four corners of the cloth in $I_b$. We assume access to a dataset of pairs of images of deformable objects, for which $n(u_a,v_a)$ is known, and a collection of correspondences and non-correspondences between each pair.
We use Blender 2.8~\cite{blender} to both generate arbitrary configurations of cloth and rope in simulation as well as to render images of these configurations for dataset curation. Blender gives us access to the underlying mesh vertices that these objects are composed of which allows us to densely sample mesh vertex pixel locations at any point.

\section{Methods}
\label{sec:methods}

\begin{figure*}[t!]
\center
\includegraphics[width=0.9\linewidth]{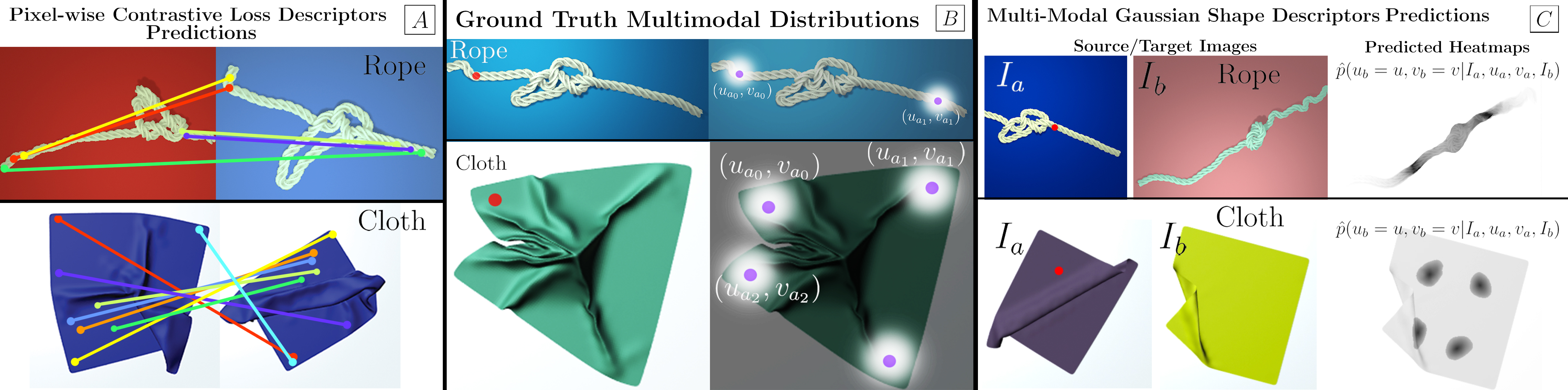}
\caption{
\small
We visualize the multi-modal ground truth distributions (B) and predicted (A,C) correspondences on domain-randomized images of both cloth and rope.  Contrastive descriptor-based methods can fail to generalize to objects with inherent symmetries, as in example A. In contrast, the predicted 2-modal and 4-modal MMGSD heatmaps for rope and cloth, respectively, appear to be sensitive to object symmetries. MMGSD exhibits the least uncertainty at object extremities and greater variance as the source pixel moves inward (C), as shown in the probability mass distributed around the rope knot (top heatmap of C).
}
\label{fig:combined}
\end{figure*}

\subsection{Preliminaries: Pixel-wise Contrastive Loss}
\label{sec:pcl}
We first review the unimodal matching method from~\cite{ganapathi2020learning,priya-rope,dense-obj-nets,visual-descriptors}. A neural network network $f$ maps $I_a$ to a $D$-dimensional descriptor volume: $f:\mathbb{R}^{W \times H \times 3} \longmapsto \mathbb{R}^{W \times H \times D}$. During training, a pair of images and sets of both matching pixels and non-matching pixels are sampled between the image pair. The following contrastive loss minimizes descriptor distance between matching pixels and pushes descriptors for non-matching pixels apart by a fixed margin $M$:

$L(I_a, I_b, u_a, v_a, u_b, v_b)$ =
\[ \begin{cases} 
     ||f(I_b)[u_b, v_b] - f(I_a)[u_a, v_a]||_2^2 $\hspace{2.5cm}$\textnormal{match}\\
     \textnormal{max}(0, M - ||f(I_b)[u_b, v_b] - f(I_a)[u_a, v_a]||_2)^2 $\hspace{0.5cm}$\textnormal{non-match}
   \end{cases}
\] 

While this method is effective at determining pixel-wise correspondences for both cloth and rope~\cite{priya-rope,ganapathi2020learning}, it does not account for inherent symmetry in these objects and therefore is susceptible to symmetric orientation based error as is shown in the 2D cloth example of Figure \ref{fig:combined}B. The authors of \cite{ganapathi2020learning} address this by limiting the rotation of the training data for a square fabric to be between $(-\dfrac{\pi}{4}, -\dfrac{\pi}{4})$, but the model still suffers from symmetric ambiguities at the boundary conditions. The authors of \cite{priya-rope} break symmetry by adding a ball to the end of the rope.



\subsection{Symmetric Pixel-wise Contrastive Loss (SPCL) Baseline}
\label{sec:symmpcl}
This method extends Section~\ref{sec:pcl} to handle multiple matches for the same source pixel $(u_a, v_a)$. Now, we try to match equivalent source pixels $\left((u_{a_i}, v_{a_i})\right)_{i=0}^n$ to a set of destination pixels $\left((u_{b_i}, v_{b_i})\right)_{i=0}^n$ that are equivalent due to symmetry by adding all pairs of pixels as matches. We use the same loss function as in Section~\ref{sec:pcl}.

While this method addresses the symmetry issue from method \ref{sec:pcl} by learning to find multiple matches in the destination image for an input pixel, we find that it is unstable and has discontinuity issues due to the contrastive nature of training. During test time, we create a heatmap of the target image by normalizing the descriptor norm differences. We then fit an $n(u_a,v_a)$-modal Gaussian distribution to the heatmap and take the $n(u_a,v_a)$ pixel modes as the predicted symmetric correspondences.

\subsection{Symmetric Distributional Loss (MMGSD)}
\label{sec:dl}
We extend a distributional descriptor network method suggested in~\cite{florence_2020} to learn an estimator $\hat{p}(u_b=u,v_b=v|I_a,u_a,v_a,I_b)$ that outputs the probability that $(u,v)$ in $I_b$ matches with $(u_a,v_a)$ in $I_a$. Specifically, we let $\hat{p}(u_b=u,v_b=v|I_a,u_a,v_a,I_b) = \frac{\exp{\|f(I_a)[u_a,v_a] - f(I_b)[u,v]\|^2_2}}{\sum_{u',v'}\exp{\|f(I_a)[u_a,v_a] - f(I_b)[u',v']\|^2_2}}$, where $f$ is a neural network with trainable parameters.
To fit $\hat{p}$, we use the cross-entropy loss function with respect to a target distribution $p$ that is an isotropic Gaussian mixture model with modes at all the ground truth pixel correspondences in $I_b$, thus accounting for all symmetric matches. For the ground truth target distributions, $\sigma$ is empirically fine-tuned to tradeoff spatial continuity in the learned distribution with overlap and collapse of modes. Using this distributional divergence loss function maintains spatial continuity between matches, and we find that this can be more stable than the method in Section~\ref{sec:symmpcl}. Additionally, predicting a distribution instead allows uncertainty estimation by computing the entropy of the predicted distribution. This method is similar to~\cite{florence_2020} but uses a multi-modal target distribution due to the multiple symmetric correspondences. As illustrated in Figure \ref{fig:combined}B and Figure \ref{fig:teaser}B, this method is successfully able to place its mass at the multiple possible matches in the target image. We fit an $n(u_a,v_a)$-modal Gaussian distribution to the predicted output distribution $\hat{p}$ and take the $n(u_a,v_a)$ pixel modes as the predicted symmetric correspondences.

\section{Quantitative Results}
\label{sec:results}
\begin{figure}[!hbtp]
\vspace{-0.25cm}
\center
\includegraphics[width=0.95\columnwidth]{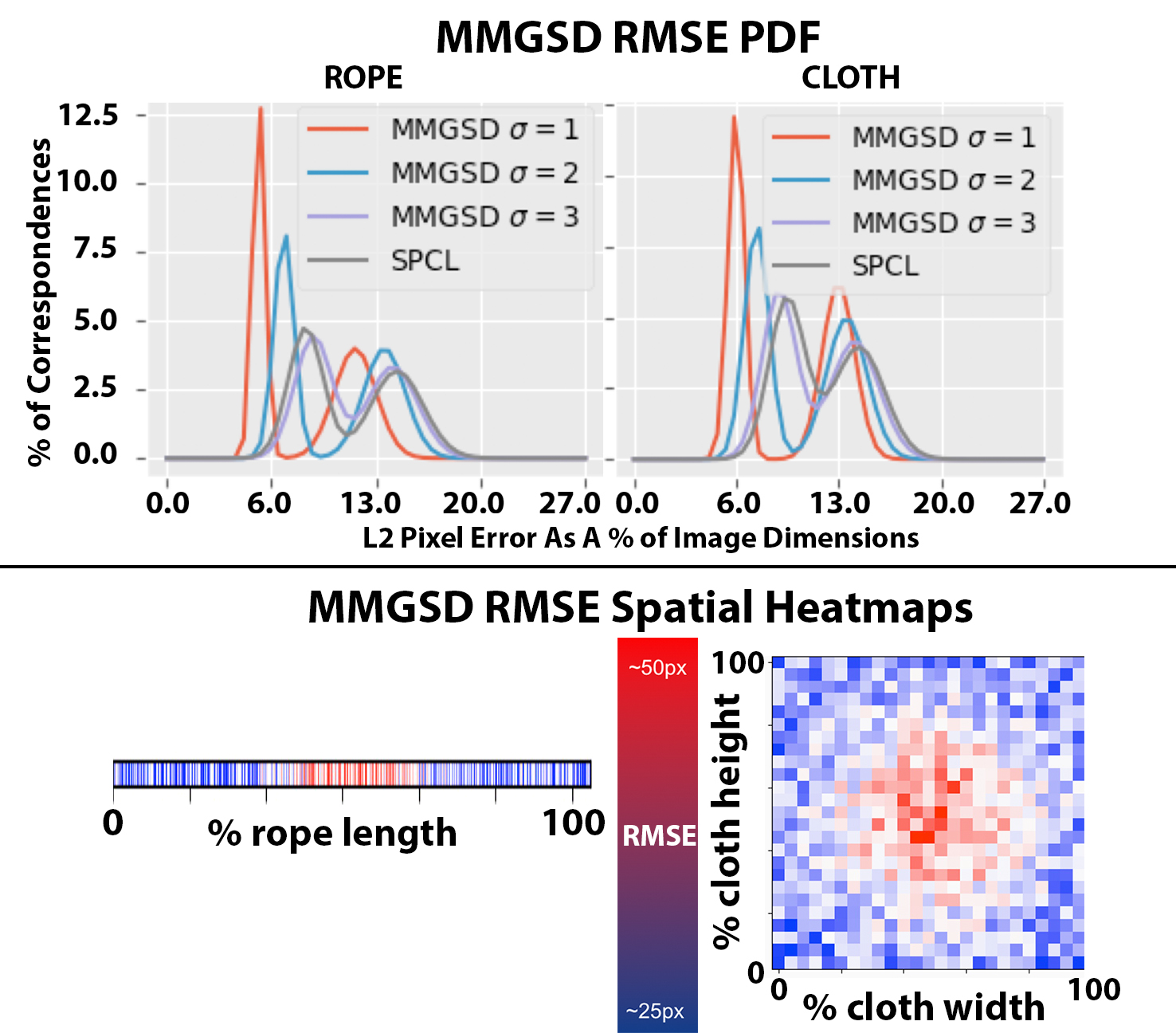}
\textbf{RMSE of Symmetric Correspondences} 
\caption{We find that MMGSD, trained with $\sigma=1$px, is able to more effectively learn symmetric correspondences over SPCL, evaluated by the PDF of correspondences with respect to L2 pixel error. For all other $\sigma$, MMGSD degrades due to intermixing of modes caused by higher variance in the ground truth target distributions. We also visualize the average RMSE in 1D and 2D space for rope and cloth, respectively, noting that MMGSD exhibits the highest error at object interiors due to modal collapse and relatively low RMSE at object extremities. This behavior of MMGSD is also suggested by the bimodal nature of the PDF with low error at object exteriors (first peak) and higher error at object interiors (second peak). }
\vspace{-0.1cm}
\label{table:rmseresults}
\end{figure}

We evaluate the quality of the symmetric learned correspondences (methods \ref{sec:symmpcl} and \ref{sec:dl}) using the root-mean-square error (RMSE) metric. Both the rope and cloth networks are trained on 3,500 training images each and evaluated on a held-out test set of 500 images. All training and testing is carried out with images of a synthetic square cloth and braided synthetic nylon rope. The cloth images are $485 \times 485$ and the rope images are $640 \times 480$ in aspect ratio. We compute the $n(u_a,v_a)$ pixel mode predictions and compare them directly to the ground truth pixel locations:
$\dfrac{1}{n(u_a,v_a)}\sum_{i=1}^{n(u_a,v_a)}||[\hat{u}_{b_i}, \hat{v}_{b_i}] - [u_{b_i}, v_{b_i}]||^2_2$ where $[u_{b_i}, v_{b_i}]$ is the ground truth pixel correspondence in $I_b$ for the source pixel $[u_{a}, v_{a}]$. We average over $625$ source pixel locations in each of $500$ test image pairs from simulation (Figure~\ref{table:rmseresults}) using a model trained on $3500$ image pairs of cloth and rope each.
    
In Figure \ref{table:rmseresults} we compare MMGSD against SPCL with the probability density function of percentage of correspondences below an L2 pixel threshold (as a percentage of the pixel dimensions of the object). We note that while MMGSD is able to predict multi-modal correspondences more effectively than SPCL, it exhibits high uncertainty and modal collapse for highly occluded regions, such as rope knots (Figure \ref{fig:combined}C), object interiors, or occluded fabric corners. This high degree of variance in the resulting heatmaps is a consequence of MMGSD attempting to preserve spatial continuity, at the expense of concentrating probability mass in isolated symmetric regions. We illustrate this in the bottom half of Figure \ref{table:rmseresults} by visualizing the source of high RMSE error on both rope and cloth. The top half of Figure \ref{table:rmseresults} also reveals this second mode centered at higher RMSE error. 


\section{Discussion and Future Work}
\label{sec:discussion}


This paper proposes an extension of dense descriptor-based pixel-wise correspondence that addresses symmetry and uncertainty estimation in deformable object tracking. In future work, we will explore generalizing MMGSD to other types of objects with task-relevant multimodal properties such as sleeves, buttons, drawstrings, or pockets on clothing. We hypothesize that the uncertainty of MMGSD --- in object interiors or in occluded parts --- would pose a challenge to tasks that involve manipulating all parts of a deformable object, such as untangling a knotted rope or performing consecutive folds on a fabric. However, we will further investigate the limitations of MMGSD and ways to utilize these measures of uncertainty while planning, such as by taking actions to disambiguate a deformable object's state. The framework presented is also applicable to rigid objects containing an axis of symmetry or multimodal properties. Additionally, we will explore learning the dynamics of these correspondences conditioned on robot action sequences. We will also explore 3D representations of deformable objects using geodesic distance as a measure of correspondence.

\vspace{-0.1cm}
\section{Acknowledgments}
\footnotesize
This research was performed at the AUTOLAB at UC Berkeley in affiliation with the Berkeley AI Research (BAIR) Lab with partial support from Toyota Research Institute. Any opinions, findings, and conclusions or recommendations expressed in this material are those of the author(s) and do not necessarily reflect the views of the sponsors. Ashwin Balakrishna is supported by an NSF GRFP and Daniel Seita is supported by an NPSC Fellowship.


\printbibliography

@STRING{icra = {{Proc. {IEEE} Int. Conf. Robotics and Automation (ICRA)}}}

@STRING{iros = {Proc. IEEE/RSJ Int. Conf. on Intelligent Robots and Systems (IROS)}}

@STRING{rss = {Proc. Robotics: Science and Systems (RSS)}}

@STRING{corl = {Conf. on Robot Learning (CoRL)}}

@article{yan2020self,
  title={Self-Supervised Learning of State Estimation for Manipulating Deformable Linear Objects},
  author={Yan, Mengyuan and Zhu, Yilin and Jin, Ning and Bohg, Jeannette},
  journal={IEEE Robotics and Automation Letters},
  volume={5},
  number={2},
  pages={2372--2379},
  year={2020},
  publisher={IEEE}
}

@inproceedings{wang2019learning,
  title={Learning robotic manipulation through visual planning and acting},
  author={Wang, Angelina and Kurutach, Thanard and Liu, Kara and Abbeel, Pieter and Tamar, Aviv},
  booktitle=rss,
  year={2019}
}

@incollection{schulman2016learning,
  title={Learning from demonstrations through the use of non-rigid registration},
  author={Schulman, John and Ho, Jonathan and Lee, Cameron and Abbeel, Pieter},
  booktitle={Robotics Research},
  pages={339--354},
  year={2016},
  publisher={Springer}
}

@inproceedings{javdani2011modeling,
  title={Modeling and perception of deformable one-dimensional objects},
  author={Javdani, Shervin and Tandon, Sameep and Tang, Jie and O'Brien, James F and Abbeel, Pieter},
  booktitle={2011 IEEE International Conference on Robotics and Automation},
  pages={1607--1614},
  year={2011},
  organization={IEEE}
}

@inproceedings{schulman2013tracking,
  title={Tracking deformable objects with point clouds},
  author={Schulman, John and Lee, Alex and Ho, Jonathan and Abbeel, Pieter},
  booktitle={2013 IEEE International Conference on Robotics and Automation},
  pages={1130--1137},
  year={2013},
  organization={IEEE}
}

@article{mcconachie2018estimating,
  title={Estimating model utility for deformable object manipulation using multiarmed bandit methods},
  author={McConachie, Dale and Berenson, Dmitry},
  journal={IEEE Transactions on Automation Science and Engineering},
  volume={15},
  number={3},
  pages={967--979},
  year={2018},
  publisher={IEEE}
}

@inproceedings{tang2017state,
  title={State estimation for deformable objects by point registration and dynamic simulation},
  author={Tang, Te and Fan, Yongxiang and Lin, Hsien-Chung and Tomizuka, Masayoshi},
  booktitle={2017 IEEE/RSJ International Conference on Intelligent Robots and Systems (IROS)},
  pages={2427--2433},
  year={2017},
  organization={IEEE}
}

@inproceedings{chi2019occlusion,
  title={Occlusion-robust deformable object tracking without physics simulation},
  author={Chi, Cheng and Berenson, Dmitry},
  booktitle={2019 IEEE/RSJ International Conference on Intelligent Robots and Systems (IROS)},
  pages={6443--6450},
  year={2019},
  organization={IEEE}
}

@article{visual-descriptors,
  title={Self-Supervised Visual Descriptor Learning for Dense Correspondence},
  author={Tanner Schmidt and Richard A. Newcombe and Dieter Fox},
  journal={IEEE Robotics and Automation Letters},
  year={2017},
  volume={2},
  pages={420-427}
}

@inproceedings{priya-rope,
  title={{Learning Rope Manipulation Policies using Dense Object Descriptors Trained on Synthetic Depth Data}},
  author={Sundaresan, Priya and Grannen, Jennifer and Thananjeyan, Brijen and Balakrishna, Ashwin and Laskey, Michael and Stone, Kevin and Gonzalez, Joseph E. and Goldberg, Ken},
  booktitle=icra,
  year=2020,
}

@inproceedings{vision-based-rope-manip,
  title={{Combining Self-Supervised Learning and Imitation for Vision-Based Rope Manipulation}},
  author={Ashvin Nair and Dian Chen and Pulkit Agrawal and Phillip Isola and Pieter Abbeel and Jitendra Malik and Sergey Levine},
  booktitle=icra,
  year={2017},
}

@inproceedings{thananjeyan2017multilateral,
  title={{Multilateral Surgical Pattern Cutting in 2D Orthotropic Gauze with Deep Reinforcement Learning Policies for Tensioning}},
  author={Thananjeyan, Brijen and Garg, Animesh and Krishnan, Sanjay and Chen, Carolyn and Miller, Lauren and Goldberg, Ken},
  booktitle=icra,
  year={2017},
}

@inproceedings{dense-obj-nets,
  title={Dense Object Nets: Learning Dense Visual Object Descriptors By and For Robotic Manipulation},
  author={Peter R. Florence and Lucas Manuelli and Russ Tedrake},
  booktitle=corl,
  year={2018}
}

@inproceedings{sim2real_deform_2018,
    title={{Sim-to-Real Reinforcement Learning for Deformable Object Manipulation}},
    author={Jan Matas and Stephen James and Andrew J. Davison},
    booktitle=corl,
    year=2018
}

@inproceedings{seita_ryan,
  title={{Deep Imitation Learning of Sequential Fabric Smoothing From an Algorithmic Supervisor}},
  author={Daniel Seita and Aditya Ganapathi and Ryan Hoque and Minho Hwang and Edward Cen and Ajay Kumar Tanwani and Ashwin Balakrishna and Brijen Thananjeyan and Jeffrey Ichnowski and Nawid Jamali and Katsu Yamane and Soshi Iba and John Canny and Ken Goldberg},
  booktitle=iros,
  year={2020}
}

@inproceedings{lerrel,
  title={{Learning to Manipulate Deformable Objects without Demonstrations}},
  author={Yilin Wu and Wilson Yan and Thanard Kurutach and Lerrel Pinto and Pieter Abbeel},
  booktitle=rss,
  year={2020},
}

@inproceedings{fabric_vsf_2020,
    author = {Ryan Hoque and Daniel Seita and Ashwin Balakrishna and Aditya Ganapathi and Ajay Tanwani and Nawid Jamali and Katsu Yamane and Soshi Iba and Ken Goldberg},
    title = {{VisuoSpatial Foresight for Multi-Step, Multi-Task Fabric Manipulation}},
    booktitle = rss,
    Year = {2020},
}

@Manual{blender,
   title = {Blender - a 3D modelling and rendering package},
   author = {Blender Online Community},
   organization = {Blender Foundation},
   address = {Stichting Blender Foundation, Amsterdam},
   year = {2018},
   url = {http://www.blender.org},
 }

@article{ganapathi2020learning,
  title={Learning to Smooth and Fold Real Fabric Using Dense Object Descriptors Trained on Synthetic Color Images},
  author={Ganapathi, Aditya and Sundaresan, Priya and Thananjeyan, Brijen and Balakrishna, Ashwin and Seita, Daniel and Grannen, Jennifer and Hwang, Minho and Hoque, Ryan and Gonzalez, Joseph E and Jamali, Nawid and others},
  journal={arXiv preprint arXiv:2003.12698},
  year={2020}
}

@inproceedings{qian2020cloth,
    title={Cloth Region Segmentation for Robust Grasp Selection},
    author={Jianing Qian and Thomas Weng and Luxin Zhang and Brian Okorn and David Held},
    year={2020},
    booktitle=iros,
}

@phdthesis{florence_2020, url={http://groups.csail.mit.edu/robotics-center/public_papers/Florence19.pdf}, journal={Dense Visual Learning for Robot Manipulation}, school={Massachusetts Institute of Technology}, author={Florence, Peter R}, year={2020}}

\clearpage

\end{document}